\documentclass[runningheads]{llncs}
\usepackage{graphicx}
\usepackage{comment}
\usepackage{tabularx}
\usepackage{adjustbox}
\usepackage{float}
\usepackage{booktabs}
\usepackage{multirow}
\usepackage{xcolor}
\usepackage{hyperref}
\usepackage[T1]{fontenc}
\begin{document}
\title{Evaluating Prompt-based Question Answering \\ for Object Prediction in the \\ Open Research Knowledge Graph\thanks{Supported by TIB Leibniz Information Centre for Science and Technology, the EU H2020 ERC project ScienceGraph (GA ID: 819536) and the BMBF project SCINEXT (GA ID: 01lS22070).}}
\titlerunning{Evaluating Prompt-based QA for the ORKG Completion}
%
\author{Jennifer D'Souza\inst{1}\orcidID{0000-0002-6616-9509} \and Moussab Hrou\inst{2} \and Sören Auer\inst{1,3}\orcidID{0000-0002-0698-2864}}
\authorrunning{D'Souza et al.}
%
\institute{TIB Leibniz Information Centre for Science and Technology, Hannover, Germany \and Gottfried Wilhelm Leibniz Universität Hannover, Germany \and L3S Research Center, Leibniz University of Hannover, Germany\\ \email{\{jennifer.dsouza,auer\}@tib.eu}}
\maketitle              
\begin{abstract}
 There have been many recent investigations into prompt-based training of transformer language models for new text genres in low-resource settings. The prompt-based training approach has been found to be effective in generalizing pre-trained or fine-tuned models for transfer to resource-scarce settings. This work, for the first time, reports results on adopting prompt-based training of transformers for \textit{scholarly knowledge graph object prediction}. The work is unique in the following two main aspects. 1) It deviates from the other works proposing entity and relation extraction pipelines for predicting objects of a scholarly knowledge graph. 2) While other works have tested the method on text genera relatively close to the general knowledge domain, we test the method for a significantly different domain, i.e. scholarly knowledge, in turn testing the linguistic, probabilistic, and factual generalizability of these large-scale transformer models. We find that (i) per expectations, transformer models when tested out-of-the-box underperform on a new domain of data, (ii) prompt-based training of the models achieve performance boosts of up to 40\% in a relaxed evaluation setting, and (iii) testing the models on a starkly different domain even with a clever training objective in a low resource setting makes evident the domain knowledge capture gap offering an empirically-verified incentive for investing more attention and resources to the scholarly domain in the context of transformer models.

\end{abstract}
\section{Introduction}

A cloze test is a language assessment where certain words (cloze text) are removed from language fragments and the participant must fill them in~\cite{taylor1953cloze}. This type of test helps language learners demonstrate certain necessary skills around language comprehension: syntax, contextual understanding, vocabulary, and factual knowledge. This philosophy of language learning has been applied to transformer language models such as BERT~\cite{kenton2019bert} and RoBERTa~\cite{liu2019roberta}. These models are pre-trained on large-scale text corpora to predict missing words or the next sentence, producing models with language comprehension abilities, around syntax, contextual understanding, vocabulary, and factual knowledge, similar to language learners.

The seminal work by Petroni et al.~\cite{petroni2019language} opened the avenue for testing these language models for their vasts store of linguistic and factual knowledge with explicit relational cloze objectives for extracting new facts from the language models for knowledge base (KB) population as a downstream task. KBs are effective solutions for accessing gold-standard relational data such as (Michael Jordan, born-in, x). The traditional method for populating such KBs with additional facts would otherwise leverage complex NLP pipelines involving entity extraction, co-reference resolution, entity linking, and relation extraction components~\cite{surdeanu2014overview} that are known to be plagued by the error propagation problem from earlier to later components in the pipeline. Instead, the powerful transformer language models as rich stores of linguistic and factual information having been pre-trained on billion-word corpus from encyclopedic sources were probed for additional facts, showing to outperform the traditional NLP pipeline method for generating relational knowledge to populate KBs as a downstream task~\cite{petroni2019language}.

While the pre-training objective prepares task-agnostic language models from large-scale corpora which were probed in the work by Petroni et al.~\cite{petroni2019language}, these language models can be further fine-tuned as an additional step with task-specific objectives which stimulates their task-specific suitability for a downstream task. To probe the transformer language models for relational facts as discussed in the earlier paragraph, the knowledge of the original models was accessed by conditioning on their latent context representations. On the other hand, to obtain a model specifically for the Question Answering (QA) downstream task, as a concrete example and the focus of this work, better versions of the models are obtained when the original models' weights are used to first initialize a task-agnostic model which is then further fine-tuned to obtain a QA task-specific model given instances of the downstream task, e.g. as defined in the Stanford Question Answering Dataset (SQuAD)~\cite{rajpurkar2016squad,rajpurkar2018know}. This work focuses on fine-tuning language models for the QA task inspired after the SQuAD dataset. The goal is to obtain a model that can extract answers from context paragraphs based on questions, and to optimize for scholarly knowledge rather than encyclopedic knowledge by tapping into the pre-trained model’s cloze task ability and the SQuAD QA fine-tuned model’s structural representation. The next two paragraphs introduce the \textit{why} and \textit{how} of our work.

\textit{Why focus on the scholarly domain?} In the face of rapid publication rates at an alarming rate of millions of articles per year~\cite{johnson2018stm}, researchers are immensely challenged in keeping up with the latest findings in scholarly publications. To address this problem, the Open Research Knowledge Graph (\href{https://orkg.org/}{ORKG})~\cite{auer2020improving}, leveraging next-generation semantic scholarly knowledge publishing tools~\cite{shotton2009semantic}, was created to make scholarly contributions more accessible and machine-actionable, enabling smart information access methods via \href{https://orkg.org/comparisons}{Comparisons}, \href{https://orkg.org/visualizations}{Visualizations}, and \href{https://orkg.org/benchmarks}{Benchmarks} thereby alleviating the researchers' knowledge comprehension problem from months or days to a matter of minutes. This growing KB of structured science-wide scholarly contributions is an unexplored resource that could be used as a testbed for discovering new facts with the help of powerful transformer language models. This could be realized in the long-run as a potential NLP service that assists ORKG users in scholarly knowledge curation and completion. Specifically, the service, directly inspired by Petroni et al's fact probing method~\cite{petroni2019language}, could be based on optimal, fine-tuned versions of the language models to discover additional objects for new incoming relations. \autoref{table-examples} shows some example instances of the proposed task. As evident from these examples, the task aims to extract objects as answers from scientific paper abstracts given a relation/predicate as a question. The task is modeled similarly to the SQuAD QA task, which extracts facts from unstructured encyclopedic knowledge using a question and answer format. We have chosen the SQuAD-based task formulation for three mains reasons: 1) the SQuAD QA task formulation intuitively transfers to the scholarly domain; 2) to exploit the vast collection of options to choose from of state-of-the-art language models fine-tuned on the QA task; and 3) to test whether these existing models' encoded statistical regularities to extract answers from a given encyclopedic knowledge context can transfer easily to the scholarly domain. We present a detailed empirical investigation for the first time of the linguistically-rich fine-tuned SQuAD QA models' capacity to transfer to a new domain, i.e. the scholarly domain, which has thus far remained unexplored.

\begin{table}[!htb]
\caption{Example instances depicting our Cloze-style adaptation of SQuAD Question Answering task as an extractive objective from the context given a question prompt. The ``Context'' column shows the relevant snippet from scholarly paper abstract given as input to the system. The ``Cloze task'' column shows the four question variants created by appending a ``Wh'' question word given an input predicate as: no label, What, Which, and How questions. The examples are selected to show the diversity in terms of the applicability of the appended question ``Wh'' question word and in terms of the various expected answer types as: a word, number, or a phrase. Text color-coded in {\color{blue}blue} show the answer extraction object from the context; text color-coded in {\color{red}red} show the appended prompt elements including ``Wh'' question words and the ``?'' symbol.}
\label{table-examples}
\scalebox{0.88}{
\begin{tabular}{p{7.5cm}lp{1.5cm}} \hline
\textbf{Context}         & \textbf{Cloze task}             & \textbf{Answer}       \\ \hline
\multirow{4}{=}{... In the following process oriented knowledge management as it was defined in the EU-project {\color{blue}PROMOTE} (IST-1999-11658) is presented and the KM-Service approach to realise process oriented ...}   & Approach name{\color{red}?} \_\_\_   & \multirow{4}{=}{{\color{blue}PROMOTE}} \\       & {\color{red}What} approach name{\color{red}?} \_\_\_        &       \\
   & {\color{red}Which} approach name{\color{red}?} \_\_\_       &          \\
   & {\color{red}How} approach name{\color{red}?} \_\_\_         &       \\ \hline
\multirow{4}{=}{... to investigate processes of community assembly contributing to biotic resistance to an introduced lineage of Phragmites australis, a model invasive species in {\color{blue}North America}. ...}              & Continent{\color{red}?} \_\_\_ & \multirow{4}{=}{{\color{blue}North America}}             \\
  & {\color{red}What} continent{\color{red}?} \_\_\_            &                    \\
  & {\color{red}Which} continent{\color{red}?} \_\_\_           &         \\
  & {\color{red}How} continent{\color{red}?} \_\_\_             &               \\ \hline
\multirow{4}{=}{... Studies done during different months (January and late February-early March) of the northeast monsoon {\color{blue}2003} revealed a fivefold increase in the average euphotic zone integrated uptake ...}     & Sampling year{\color{red}?} \_\_\_  & \multirow{4}{=}{{\color{blue}2003}}      \\
    & {\color{red}What} sampling year{\color{red}?} \_\_\_        &     \\
  & {\color{red}Which} sampling year{\color{red}?} \_\_\_       &        \\
  & {\color{red}How} sampling year{\color{red}?} \_\_\_         &         \\ \hline
\multirow{4}{=}{{\color{blue}Solid lipid nanoparticles} (SLNs) are nanocarriers developed as substitute colloidal drug delivery systems parallel to liposomes, lipid emulsions, polymeric nanoparticles, and so forth. ...}       & Type of nanocarrier{\color{red}?} \_\_\_       & \multirow{4}{=}{{\color{blue}Solid lipid nanoparticles}} \\
   & {\color{red}What} type of nanocarrier{\color{red}?} \_\_\_  &     \\
   & {\color{red}Which} type of nanocarrier{\color{red}?} \_\_\_ &      \\
  & {\color{red}How} type of nanocarrier{\color{red}?} \_\_\_   &  \\    \hline   
\end{tabular}}
\end{table}

\textit{How to obtain an optimal model for a new domain?} To pre-train or fine-tune a language model for QA on a new domain, the traditional method is to use expensive human-labeled data. Instead, inspired from prior work~\cite{fabbri2020template,schick2021exploiting,zhong2022proqa}, this study uses two strategies: 1) template-based unsupervised generation of structured data similar to SQuAD QA data from the ORKG KB, and 2) structural prompt-based learning over state-of-the-art SQuAD-specific fine-tuned transformer models for the scholarly domain.
\textit{How to obtain an optimal model on a new domain?} The similar structured QA data can serve as prompts to stimulate the knowledge acquired in the fine-tuned language models to model knowledge generalizations over the new domain based on the unified input schema in turn resulting in an optimal trained model in a resource-scarce setting. The idea is to apply a ``Wh'' question template on an ORKG structured contribution predicate, and then generate a question based on that predicate whose answer is the ORKG object that can be found as a contiguous span in the paper's abstract. Having created a question for all (context, answer) pairs, we then reserve a portion of the data for prompt-based fine-tuning of variants of state-of-the-art SQuAD transformer models on this data and evaluate on the remaining held-out version of the synthetically generated ORKG QA data.

Summarily, our contributions are: 1) we introduce a template-based unsupervised question generation framework for the scholarly domain, similar in format to the SQuAD dataset; 2) we report, for the first time, a detailed empirical analysis of the scholarly domain object prediction task using prompt-based QA task and state-of-the-art transformer models as rich stores of linguistic, probabilistic, and factual parameters, thereby testing the transferability of these pre-trained models on a novel domain. Our \href{https://data.uni-hannover.de/dataset/evaluating-squad-based-question-answering-for-the-open-research-knowledge-graph-completion}{dataset}, \href{https://github.com/as18cia/thesis_work}{code}, and \href{https://huggingface.co/Moussab}{models} are publicly released.





\section{Task Definition}

The article introduces a new task of object extraction for RDF knowledge graph statements, which is structurally formulated based on the SQuAD QA task dataset~\cite{rajpurkar2016squad,rajpurkar2018know}. The task is an extractive task in a machine reading comprehension setting, where the model is expected to extract the answer as a contiguous span to a given question. Our task is: given an ORKG \textit{predicate} formulated as a question using an unsupervised template-based generation prompting function $f_{prompt}(x)$, to extract the \textit{object} answer from the corresponding paper \textit{Abstract} context. For unsupervised question generation as $f_{prompt}(x)$, a static template pattern is defined. Inspired by the prior seminal work on this theme~\cite{fabbri2020template}, the $f_{prompt}(x)$ template is: ``Wh''+predicate+``?'', where ``Wh'' $\in$ \{What, Which, How\} as the most common question types. Each ``Wh'' question results in a question-type-specific homogeneous dataset where the same prompt template considering the specific question-type word was applied to all predicates. Finally, a note on the expected object answer granularity considered in this task: They include three different granularities: tokens, span as a short multi-token phrase, and sentences.

\section{\textsc{Prompt}-ORKG: Our Scholarly Knowledge Question Answering Corpus}
\label{corpus}

We aim to create a high-quality corpus of questions to prompt language models to extract the correct answer from a context-question pair. For this, we adopt the structural format of the SQuAD dataset to prompt already fine-tuned language models and aim to stimulate the models in the most optimal setting to demonstrate their transferability on a new domain.


\begin{figure}[!htb]
\includegraphics[width=\textwidth]{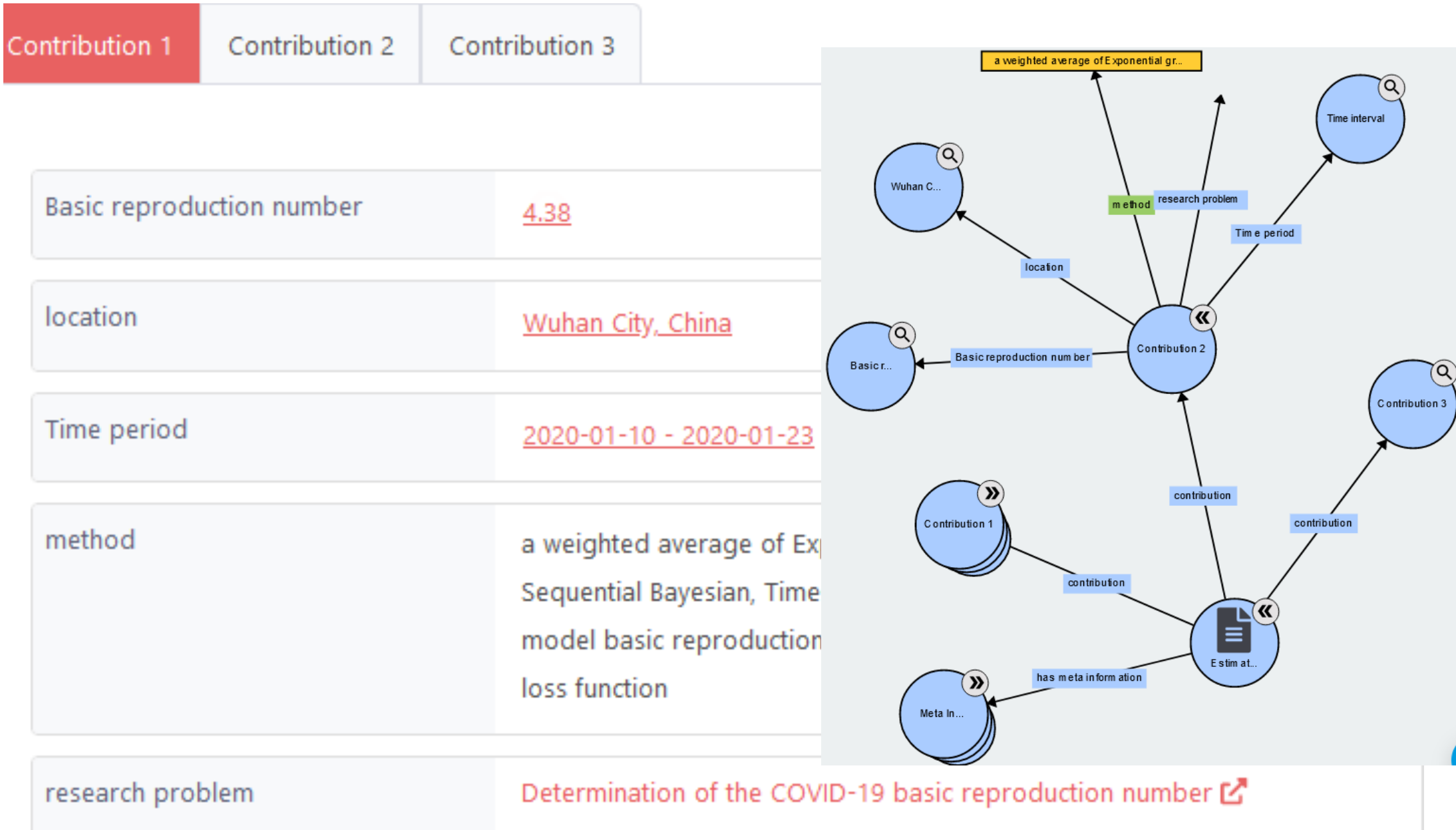}
\caption{A scholarly paper structured contribution view in the Open Research Knowledge Graph with the backend graph structure overlay on the frontend interface. The original paper can be accessed here \url{https://orkg.org/paper/R44743/}} \label{fig-orkg-paper}
\end{figure}

\subsection{The Open Research Knowledge Graph as a Knowledge Base}

The ORKG semantic scholarly publishing model represents \textit{scholarly contributions} in a structured and machine-actionable way. Scholarly contributions refer to the specific content of a scholarly article that highlights its central findings~\cite{ncg}. In general, a contribution describes the research problem addressed and has salient properties that need to be modeled with a generalization objective so that it can be compared with other articles addressing the same research problem.~\cite{oelen2020generate}. E.g., the predicates \textit{basic reproduction number}, \textit{confidence interval (95\%)}, \textit{location}, and \textit{time period} are used to describe Covid-19 reproductive number estimates across various epidemiology articles (\url{https://www.orkg.org/orkg/comparison/R44930}). \autoref{fig-orkg-paper} illustrates a structured contribution discussed in this example on the ORKG platform. 

The ORKG semantic web publishing model uses a set of contribution triples to describe a paper in a structured format using RDF subject-predicate-object triples statement paradigm. Users can add new properties or attributes at any time using the ORKG frontend interface. Reusing our earlier running example, some of its triples would be: (Contribution, location, Wuhan City, China), (Contribution, basic reproduction number, 4.38), (Contribution, Time period, 2020-01-10 - 2020-01-23).

In the context of the ORKG data explained, we follow a sequence of steps to tailor the KB as \textit{scholarly knowledge object prediction} formulated as extractive QA structured per the SQuAD dataset. To satisfy this task objective, with the help of the \href{http://tibhannover.gitlab.io/orkg/orkg-backend/api-doc/}{ORKG Rest API} we queried only portions of the graph to extract specifically all triples in the ORKG KB that hinged on the Contribution node as the subject. Now our challenge is to obtain the (context, question, answer) data instances. First, we address our creation of the preliminary data for the question-and-answer units. As preliminary data, for the question unit, the predicates in each triple were chosen as candidates, and, for the answer unit, their corresponding object nodes were selected. Some statistics of this initial raw dataset are depicted in the `Before' column of \autoref{table-data-stats}. Notably, in our data, the number of contributions is greater than the number of papers since a paper can have more than one contribution. Also the (predicate, object) pairs are relatively sparse spanning over 200 different research fields in the \href{https://orkg.org/fields}{ORKG research fields taxonomy}.

\begin{table}[!htb]
\centering
  \caption{Our scholarly knowledge question answering for object prediction task dataset statistics on the raw data (`Before' column) and after data cleaning (`After' column).}
  \label{table-data-stats}
  \begin{tabular}{|l|r|r|r|}
      \toprule
       \bf Statistic parameter & \bf Before & \bf After\\
      \midrule
      number of unique papers       & 9,379  & 2,710  \\
      number of unique contributions & 14,499 & 3,059 \\      
      number of (predicate,object) pairs  & 116,421 & 5,909 \\
      number of unique predicate labels & 3,436 & 853 \\
      number of unique object labels & 38,234 & 3,524 \\
      avg. number of tokens per predicate label   & - & 2.01    \\
      avg. number of tokens per object label & - & 2.43    \\
      number of unique abstracts & - & 2649    \\
      avg. number of tokens per paper abstract  & - & 196.97       \\
      number of abstracts with more than 510 tokens & - & 37            \\
      number of unique abstracts with more than 510 tokens & - & 14      \\
      \bottomrule
    \end{tabular}
\end{table}

\subsection{Template-based Questions Generation}

This section describes how the \textit{question} unit is created of the needed (context, question, answer) data instance. Based on the predicate, we generate four variants of our dataset based on the $f_{prompt}$ function: 1) 3 variants with the 3 most common ``Wh'' question words (i.e. What, Which, and How) which are generated in an unsupervized manner following the $f_{prompt}$ template ``Wh+predicate+?,'' and 2) 1 variant as a cloze-style question but instead of using the [MASK] token, we use the ``?'' symbol following the template ``predicate+?.'' While, our experimental settings are defined on these four corpus variants based on the generated question unit, we additionally evaluate a dataset called ``unchanged.'' In this dataset, the predicate is posed as a question unit as it is with neither the ``?'' symbol or question words appended, nor case changes. Thus we can contrast evaluation results with the ``unchanged'' dataset to show the effectiveness of adopting the specially designed SQuAD-format prompt-based QA setting.

\subsection{Obtaining a SQuAD-format Corpus for the Scholarly Domain}


Having addressed the question unit, this section explains how the SQuAD (context, question, answer) structured corpus is generated by addressing the remaining units starting with the context. Unlike the SQuAD dataset, in our corpus, the context is posited as the paper Abstract. The \href{https://gitlab.com/TIBHannover/orkg/orkg-abstracts}{ORKG abstracts fetching script} was used to query external metadata services such as \href{https://api.crossref.org/}{Crossref} or \href{https://api.semanticscholar.org/}{SemanticScholar} with the paper DOI or title (when DOI is absent) to create an abstracts corpus to serve as the context unit. However, not all papers in the ORKG get abstracts from these external services. Specifically, we were able to obtain abstracts for 5,486 (58,5\%) out of the 9,379 papers in the raw data. This results in pruning all the rows in our triples dataset whose abstracts could not be fetched.

Next, we focus on the \textit{answer} unit i.e. from all object candidates selecting those that are suitable for our extractive QA task. A precondition for selecting suitable objects was to ensure that they can be found in the paper Abstract. Thus, the 116,421 (predicate, object) pairs in the raw dataset were reduced to 14,499 (predicate, object) pairs considering papers whose abstract could be fetched and where the object label was found in the abstract. The resulting data was then cleaned via a two-step methodology: 1) deduplication, i.e. all rows with the same (predicate, object) and context were dropped; and 2) a heuristics-based method to clean the dataset of object labels that were deemed as unsuitable extraction targets. These included whole numbers from 0 to 999, the hyphen symbol, alphabets, boolean values as T/F/yes/no, not applicable string i.e. na, stop words such as all/and/or etc., and a collection of non-informative phrases like ``any track'' or ``method'' etc. Finally, our dataset comprised 5,909 (predicate, object) pairs coupled with a context (see row 3 in \autoref{table-data-stats}) which we name the \textsc{Prompt}-ORKG corpus.

\subsubsection{Object categorization} The final step to obtain a SQuAD-format corpus was to carry out a coarse-grained categorization of the objects. The object types in our dataset while inspired by the SQuAD 10 syntactic types~\cite{rajpurkar2016squad} were finally chosen to reflect the peculiarities of our data. They range between generic syntactic types (e.g., noun, adjective, noun phrase, sentence) and domain-specific types (e.g., research problem, url, count/measurement). \autoref{dataset-object-categories} shows the types and their coverage with a supporting example for clarity. The object typing was carried out as described below.

    \begin{itemize}
        \item research problem: any object label in a row with the predicate label ``has research problem.''
        \item url: any object label that starts with ``http.''
        \item location: any object label with predicate label in [country, city, location, continent, has location, study location, countries].
        \item year/date: any whole number between 1000 and 2100.
        \item number: any object label consisting of only digits after removing the minus symbol (-), dot (.), and comma (,).
        \item count/measurements: any object label that contains at least a number in addition to another string(s). E.g., ``5 meters.''
        \item noun: \href{https://pypi.org/project/spacy/}{spaCy} was leveraged to determine the noun part-of-speech for a token.
        \item adjective: spaCy classified adjective part-of-speech.
        \item acronym: token with one upper case alphabet.
        \item noun phrase: objects with between 2 and 5 tokens and the last token is a noun.
        \item adjective phrase: objects with between 2 and 5 tokens and the last token is an adjective.
        \item sentence: any object with more than 4 tokens that does not fall under any other category.
    \end{itemize}

To ensure the objects were typed as only one category, the categorization heuristics were precedence-ordered following the listing above. Thus an object was typed per the earliest heuristic that applied to it. From \autoref{dataset-object-categories}, we can see that with a 58.9\% coverage, nouns or noun phrases were the most common. 

\begin{table}[!htb]
\centering
  \caption{The heuristically assigned categories for the objects in \textsc{Prompt}-ORKG.}
  \label{dataset-object-categories}
  \begin{tabular}{|l|l|c|l|}
      \toprule
       & \bf Object category & \bf \% Coverage & \bf Example\\
      \midrule
      1 & Noun              & 29.85         & Transistors  \\
      2 & Noun phrase       & 28.74         & data mining  \\
      3 & Acronym           & 10.12         & HMM \\
      4 & Research problem  & 9.53          & Performance of thin-film transistors \\
      5 & Adjective               & 4.72          & high \\
      6 & Location          & 4.37          & Serbia \\
      7 & Number            & 3.93          & 4977 \\
      8 & Count/measurement & 3.47          & 2.45 GHz \\
      9 & Sentence          & 2.76          & raw data dumps and HDT files \\
      10 & Year/date         & 1.95          & 2011 \\
      11 & URL               & 0.30           & https://github.com/giannisnik/mpad  \\
      12 & Adjective phrase  & 0.27          & Unsupervised and Adaptive \\
      \bottomrule
    \end{tabular}
\end{table}

\section{Models}

With the \textsc{Prompt}-ORKG dataset for the scholarly domain in place, we relegated attention to selecting three optimal transformer model variants to test as machine learners on our newly introduced, previously unexplored problem setting. \textit{The machine learning test specifically sought out empirical evidence for the transferability of the probabilistic parameters of the existing large-scale transformer models.} In this respect, we were interested in two main strengths of the transformer models: to query over an open class of relations~\cite{petroni2019language} and the ability to train them on structurally similar data~\cite{fabbri2020template,schick2021exploiting}. Thus selecting optimal SQuAD transformer model variants were a natural choice since the structural patterns in the \textsc{Prompt}-ORKG dataset emulate SQuAD.

\subsection{Three Optimal SQuAD Transformer Model Variants}

The selected language model variants are based on BERT~\cite{bert} which seminally introduced the ``masked language model'' (MLM) cloze-based pre-training objective based on a bidirectional self-attention architecture~\cite{vaswani2017attention}. The traditional setup includes two steps: pre-training on a large-scale corpus with a cloze objective for language comprehension, and fine-tuning the initialized pre-trained model parameters using labeled data from the SQuAD QA downstream task.



\textbf{BERT pretrained, SQuAD2.0 finetuned} For the first language model, we sought out the state-of-the-art BERT-based variant from HuggingFace. This model is called \href{https://huggingface.co/deepset/bert-base-cased-squad2}{deepset/bert-base-cased-squad2}. It leverages the standard BERT-base cased, initialized model parameters that were pre-trained on the BookCorpus \cite{bookcorpus} plus English Wikipedia; these were then further fine-tuned using the SQuAD2.0 corpus~\cite{rajpurkar2018know}. This model reports 74.67\% F1-score on SQuAD2.0.

\textbf{RoBERTa pretrained, SQuAD2.0 finetuned} The second language model variant was chosen based on RoBERTa~\cite{roberta}. RoBERTa transformer model pre-training improves upon BERT in several aspects: (1) improved BERT design choices and training strategies using longer training time, bigger batches, longer sequences, dynamic masking patterns, and removing the next sentence prediction objective; and (2) increased underlying training data size to include BookCorpus \cite{bookcorpus}, \href{http://web.archive.org/save/http://commoncrawl.org/2016/10/newsdataset-available}{CommonCrawl News}, \href{http://web.archive.org/save/http://Skylion007.github.io/OpenWebTextCorpus}{OpenWebText} from Reddit, and Stories \cite{trinh2018simple}. The selected HuggingFace model is called \href{https://huggingface.co/deepset/roberta-base-squad2}{deepset/roberta-base-squad2}. It was further fine-tuned on SQuAD2.0, and reports 82.91\% F1-score.

\textbf{MiniLM pretraining distillation, SQuAD2.0 finetuned} The third variant is selected from the line of work that focused on compressing large pre-trained models into small and fast pre-trained models while achieving competitive performance with the specific objective of serving in practical applications. From these models, generically referred to as distilled variants of the base pre-trained models such as BERT, we select the outperforming MiniLM \cite{minilim} variant. The selected model after reviewing the HuggingFace collection is called \href{https://huggingface.co/deepset/minilm-uncased-squad2}{deepset/minilm-uncased-squad2} which was fine-tuned similar to the prior two variants on SQuAD2.0. It achieves 79.49\% F1-score.

Each of these respective models has been fine-tuned over massive amounts, ranging on 100,000+ instances, of SQuAD QA data. Over these model variants, this work investigates an additional step of \textit{prompt-based training} that has been shown to optimally stimulate a fine-tuned model's probabilistic parameters to generalize in a resource-scarce setting~\cite{liu2021pre} that we defined as the \textsc{Prompt}-ORKG corpus.  


\section{Results and Discussion}

\subsection{Experimental Setup}

\textbf{Dataset.} \textsc{Prompt}-ORKG presents a few-shot learning scenario. On this corpus, we need to obtain relevant training and evaluation dataset splits. The dataset was split based on the predicate labels as follows. For predicate labels with 10 or more instances, 75\% of the corresponding data instances are assigned to the training data and the rest are reserved as the evaluation set. The remaining data snapshot based on predicate labels with less than 10 instances is simply relegated as the training set. Finally, for each of the four templated dataset variants and the fifth ``unchanged'' variant discussed in \autoref{corpus}, the training dataset consists of 4745 data points (82\% of the entire dataset) while the evaluation set consists of 1036 data points (18\%).

\noindent{\textbf{Hyper-parameter tuning.}} All models are tuned for 4 epochs, learning rate $\in$ \{0.0001, 0.00005\}, train batch size = 8, eval batch size = 8, and weight decay = 0.01. At the end of the training phase of each experiment, only the best model is saved and used in the evaluation phase.

\noindent{\textbf{Metrics.}} Evaluations are considered in two main settings: 1. strict, i.e. exact match; and 2. relaxed, i.e. containment match where the gold answer is checked to be contained in the predicted answer. In both settings, the main metric is F1 score and secondarily accuracy is also applied. Note that after prediction, the answers undergo minimal post-processing to be suitable for evaluation. This entailed trimming the trailing and leading white spaces, converting all answers to lower case, and removing the following special characters: ., comma, ;, :, -, ), (, \_ and +, if they are at the end of the answers.


\subsection{Evaluations}
In this section, we present the results and discuss observations from each of our 30 total experiments considering 5 dataset variants, 3 optimal SQuAD QA models, and 2 experimental settings i.e. vanilla and after prompt-based learning. The core experimental results are depicted in Tables \ref{f1score-exact} and \ref{f1score-inexact} in terms of F1-scores, and parenthesized accuracies, in the exact-match vs. relaxed settings, respectively. We discuss the experimental results with respect to three main research questions.

\textbf{RQ1: What is the impact of the \textsc{Prompt}-ORKG SQuAD-format structural QA task formulation on the transferability of the large-scale SQuAD fine-tuned language models?} This question subsumes two sub-questions. RQ1.1: How do the BERT model performances contrast when queried out-of-the-box, i.e. as vanilla models, versus after being trained on our corpus variants? RQ1.2: Is it effective to adopt the SQuAD-format structural representation to optimally query the large-scale SQuAD fine-tuned language models? For RQ1.1, examining both F1 scores and accuracies, in the exact match setting (\autoref{f1score-exact}), the results from the trained models on our corpus variants are approximately 25\% (and 35\%) higher, respectively, than when testing the untrained ``vanilla'' models on data from a new domain. For RQ1.2, examining the *row avg* exact match F1 scores in \autoref{f1score-exact}, we see the results from the SQuAD-format dataset variants are 5\% or 6\% higher than the ``unchanged'' dataset. Note the SQuAD-format dataset variants report results in the range 21\% to 22\% F1 scores, while the ``unchanged'' dataset evaluations report only 16\% F1 score. Thus we can conclude that indeed the SQuAD-format task formulation as the four variants in the \textsc{Prompt}-ORKG corpus is an effective strategy to optimally stimulate the probabilistic parameters of the SQuAD fine-tuned language models towards their transferability on a new domain, i.e. the scholarly domain.

Thus our findings cumulatively are as follows. Training the models on our corpora produce more effective predictors compared to the vanilla models with performances as low as 1\% reflecting the domain gap of the language models between the generic domain and the scholarly domain. To obtain optimal versions of the SQuAD fine-tuned language models, formulating the task dataset in SQuAD format is an effective strategy to obtain optimally trained models. 

\textbf{RQ2: Which dataset variant produced the most optimal models?} Given that we have 2 evaluation tables, the F1 scores in the exact match setting shown in \autoref{f1score-exact} were considered the reference evaluations on which to base conclusions. Examining the *row avg* results in the Table, we observed that the results from the \textit{which} variant of the \textsc{Prompt}-ORKG corpus was statistically insignificantly better than the \textit{what} variant with 22.6\% F1 score versus 22.2\% F1 score. This indicates directions for future work to examine the automated generation of the questions by selecting and applying the most suitable question type \textit{which} or \textit{what} for the predicates. Across all four tables, examining the *column avg* results, the RoBERTa SQuAD model proved optimal for transferability to a new domain.

\begin{table}[!htb]
  \caption{F1-score (and parenthesized Accuracy) results in the exact-match setting over the 4 dataset variants from the 3 models with cell values as ``vanilla models''/``after prompt-based training.''}
  \label{f1score-exact}
  \centering
    \begin{tabular}{|c|c|c|c|c|} 
      \toprule
      \bf \stackbox[c]{Dataset \\variant} & \bf \stackbox[c]{bert-base-cased\\-squad2} & \bf \stackbox[c]{roberta-base\\-squad2} & \bf \stackbox[c]{minilm-uncased\\-squad2} & \bf *row avg*\\
      \midrule
      unchanged     &0.5/11.2 (1.0/29.4) & 2.5/22.7 (1.8/35.5) & 0.5/16.2 (0.4/35.5) &\colorbox{lightgray}{1.2/16.7 (1.1/33.5)} \\       
      none          &1.4/23.5 (1.4/31.7) & 4.0/23.0 (2.5/37.5) & 3.6/18.0 (3.5/35.4) &\colorbox{lightgray}{1.8/21.5 (2.5/34.9)}  \\
      what          &1.5/25.7 (1.0/33.8) & 5.1/21.2 (4.3/35.9) & 6.3/19.8 (5.1/36.1) &\colorbox{lightgray}{4.3/22.2 (3.5/35.3)}  \\
      how           &0.3/19.8 (0.5/34.0) & 3.0/20.9 (2.2/36.4) & 4.6/22.3 (4.0/33.8) &\colorbox{lightgray}{2.6/21.0 (2.2/34.7)}  \\
      which         &1.9/17.8 (1.6/33.5) & 5.5/24.0 (4.5/36.6) & 5.9/25.9 (5.5/36.8) &\colorbox{lightgray}{4.4/22.6 (3.9/35.6)}  \\ \hline
      *column avg*    &\colorbox{lightgray}{1.1/19.6 (1.1/32.5)} &\colorbox{lightgray}{4.0/22.4 (3.1/36.4)} &\colorbox{lightgray} {4.2/20.4 (3.7/35.5)}  &\colorbox{lightgray}{-}             \\
      \bottomrule
    \end{tabular}
\end{table}

\begin{table}[!htb]
  \caption{F1-score (and parenthesized Accuracy) results in the relaxed setting over the 4 dataset variants from the 3 models with cell values as ``vanilla models''/``after prompt-based training.''}
  \label{f1score-inexact}
  \centering
    \begin{tabular}{|c|c|c|c|c|}
      \toprule
      \bf \stackbox[c]{Dataset \\variant} & \bf \stackbox[c]{bert-base-cased\\-squad2} & \bf \stackbox[c]{roberta-base\\-squad2} & \bf \stackbox[c]{minilm-uncased\\-squad2} & \bf *row avg*\\
      \midrule
      unchanged     & 6.7/17.1 (9.2/42.9) & 7.6/23.7 (5.7/49.7) & 18.6/22.9 (18.2/47.5) &\colorbox{lightgray}{11.0/21.2 (11.0/46.7)}  \\     
      none          & 6.2/31.7 (5.9/43.4) & 20.3/32.3 (14.8/50.6) & 21.3/26.1 (16.0/48.3) &\colorbox{lightgray}{15.9/30.0 (12.2/47.4)}  \\
      what          & 7.3/34.6 (5.6/46.0) & 22.8/31.0 (17.0/49.5) & 24.4/26.2 (16.5/46.0) &\colorbox{lightgray}{18.2/30.6 (13.0/47.2)}  \\
      how           & 6.8/27.6 (8.2/45.2) & 23.7/30.4 (16.5/51.2) & 22.7/28.5 (16.3/47.0) &\colorbox{lightgray}{17.7/28.8 (13.7/47.8)}  \\
      which         & 8.3/24.0 (7.4/43.9) & 25.1/35.9 (18.2/48.6) & 23.0/36.4 (18.0/47.9) &\colorbox{lightgray}{18.8/32.1 (14.5/46.8)}  \\ \hline
      *column avg*    &\colorbox{lightgray}{7.1/27.0 (7.2/44.3)}   &\colorbox{lightgray}{19.9/30.7 (14.4/49.9)} &\colorbox{lightgray} {22.0/28.0 (17.0/47.3)}  &\colorbox{lightgray}{-}             \\
      \bottomrule
    \end{tabular}
\end{table}

\textbf{RQ3: Which of the \textsc{Prompt}-ORKG corpus object types was easiest/challenging to predict?} Examining the results in \autoref{accuracy-object-category}, we see that the Location, Adjective, Noun, Acronym, Research problem, and Year/date object types were easiest to predict with accuracies above 40\% with values 53.3\%, 50\%, 44.6\%, 43.5\%, 43.4\%, and 42.9\%, respectively. The Sentence type could not be predicted at all by any of the models and thus presented the most challenging type. This is somewhat an expected result since the fine-tuned language models were trained only over object types of a few tokens thus base model parameters were lacking to generalize over object types with many more tokens. The low performance on the noun phrase object type can be similarly attributed. \autoref{avg-predicted-tokens} offers some insights into the average number of tokens in our corpus versus the average tokens from the model predictions obtained. Low performance on the URL object type can be attributed to its under-representation in our corpus. 

\begin{table}[!tb]
  \caption{Fine-grained accuracy results for the 12 object categories in the exact-match/relaxed settings for best model (RoBERTa) and dataset variant (\textit{which}) combination.}
  \label{accuracy-object-category}
  \centering
 \begin{tabular}{|l|r|r|}
      \toprule
       \bf Object category & \bf \stackbox[c]{\% Coverage in the \\ evaluation dataset}& \bf \% Accuracy \\
      \midrule
      Sentence            & 3        & 0.0/3.3  \\
      Count/measurement   & 3        & 7.1/14.3 \\
      Number              & 0.5      & 20/40.0 \\
      Noun phrase         & 30       & 25.8/37.4 \\
      URL                 & 0.25     & 33.3/33.3 \\      
      Year/date           & 1        & 42.9/57.1 \\
      Research problem    & 14       & 43.4/53.8 \\      
      Acronym             & 9        & 43.5/59.8 \\
      Noun                & 30       & 44.6/58.7 \\
      Adjective           & 3        & 50/57.1 \\
      Location            & 6        & 53.3/57.4 \\
      \bottomrule
    \end{tabular}
\end{table}

\begin{table}[!htb]
\centering
  \caption{Average number of tokens for gold answers and for the predicted answers from the vanilla/fine-tuned models on the most effective prompt-based dataset variant.}
  \label{avg-predicted-tokens}
 \begin{tabular}{|l|r|r|}
      \toprule
       \bf Model & \bf \stackbox[c]{Gold Avg. \\ \# tokens}& \bf \stackbox[c]{Predicted Avg. \\ \# tokens}\\
      \midrule
      bert-base-cased-squad2      & 2.4        & 4/7.3  \\
      roberta-base-squad2          & 2.4        & 9.3/5.5 \\
      minilm-uncased-squad2        & 2.4        & 17/5.7 \\
      \bottomrule
    \end{tabular}
\end{table}

\vspace{-1em}
\subsection{Discussion}

In this section, we dive deeper into the results from the relaxed match settings, where the predictions were simply checked to contain the gold answer, shown in \autoref{f1score-inexact} to offer insights into whether our formulated QA task over the scholarly domain is promising for further explorations as future work. From \autoref{f1score-inexact}, we see that prompt-training the language models in a few shot learning setting over data from the new scholarly domain still achieves F1 scores close to or above 35\% in the optimal settings and accuracies above 45\% in most settings and 50\% in some settings. 
Given this, we can conclude that the language models show promise of transferring effectively to the new scholarly domain provided the underlying task dataset formulation is suitably made incorporating the original language models task dataset formulation. Thus we elicit specific areas of further exploration from our present study as the scope for future work as follows. 1) The creation of a heterogeneous corpus with questions formulated based on the automated selection of suitable question words depending on the predicate or the answer. 2) Testing a different prompting method for the language models, i.e. the instructional prompting method that proved very effective for the language models when tested in various zero-shot learning scenarios~\cite{wei2021finetuned}. Furthermore, with the instructional prompt writing or task dataset formulation, the collection of pre-trained language models to experiment with is even broader without being restricted to the SQuAD fine-tuned models as in our present work. 3) This work has released uniform conclusive experimental results for the following experimental setup: commonsense pretraining domain + commonsense QA finetuning domain from SQuAD + Prompt-ORKG scholarly finetuning domain. The hypothesis then tested was these commonsense-trained-tuned models parameter generalizability to the scholarly domain. We test this hypothesis because we noted that the commonsense domain receives a lot of attention in the release of annotated data for various machine learning tasks. On the other hand, the scholarly domain–not so much. So we want to advocate for the use of these commonsense models in the best transfer learning settings. Nevertheless, given the availability of scholarly domain pretrained language models like SciBERT~\cite{beltagy2019scibert} or BioBERT~\cite{lee2020biobert}, their systematic investigations in future work is recommended, where inspired from this work, the experimental setup would be: scholarly pretraining domain + commonsense QA finetuning domain from SQuAD + Prompt-ORKG scholarly finetuning domain.

\section{Conclusions}

The prompt-based learning paradigm~\cite{liu2021pre,wei2022chain} is increasingly used in NLP for fine-tuning BERT-based QA models on domains with less training data. In this work, we demonstrated the applicability of SQuAD-based prompt format training of BERT models on scholarly data for object prediction in the ORKG. Our experiments showed promising results in terms of domain transferability of the models with the right training strategy, however, the model performances reflect a vast scope for further improvement.

%
%
%
\bibliographystyle{splncs04}
\bibliography{mybib}

\end{document}